\documentclass[a4paper, 10pt, conference]{ieeeconf}       
\IEEEoverridecommandlockouts                              
\usepackage{graphics} 
\usepackage{epsfig} 
\usepackage{multirow}
\usepackage{color}
\usepackage{comment}
\usepackage{afterpage}
\usepackage{setspace}
\usepackage{amsmath}
\usepackage{algorithm}
\usepackage[noend]{algorithmic}

\title{\LARGE \bf
Online Self-Supervised Learning for Object Picking:\\
Detecting Optimum Grasping Position using a Metric Learning Approach
}

\author{Kanata Suzuki, Yasuto Yokota, Yuzi Kanazawa  and Tomoyoshi Takebayashi
\thanks{All authors are associated with Fujitsu Laboratories LTD., Kanagawa 211-8588, Japan. E-mail: {\tt\small suzuki.kanatai@fujitsu.com}}
}

\begin{document}
\maketitle
\thispagestyle{empty}
\pagestyle{empty}

\begin{abstract}
Online self-supervised learning methods are attractive candidates for automatic object picking. Self-supervised learning collects training data online during the learning process. However, the trial samples lack the complete ground truth because the observable parts of the agent are limited. That is, the information contained in the trial samples is often insufficient to learn the specific grasping position of each object. Consequently, the training falls into a local solution, and the grasp positions learned by the robot are independent of the state of the object. In this study, the optimal grasping position of an individual object is determined from the grasping score, defined as the distance in the feature space obtained using metric learning. The closeness of the solution to the pre-designed optimal grasping position was evaluated in trials. The proposed method incorporates two types of feedback control: one feedback enlarges the grasping score when the grasping position approaches the optimum; the other reduces the negative feedback of the potential grasping positions among the grasping candidates. The proposed online self-supervised learning method employs two deep neural networks. : a single shot multibox detector (SSD) that detects the grasping position of an object, and Siamese networks (SNs) that evaluate the trial sample using the similarity of two input data in the feature space. Our method embeds the relation of each grasping position as feature vectors by training the trial samples and a few pre-samples indicating the optimum grasping position. By incorporating the grasping score based on the feature space of SNs into the SSD training process, the method preferentially trains the optimum grasping position. In the experiment, the proposed method achieved a higher success rate than the baseline method using simple teaching signals. And the grasping scores in the feature space of the SNs accurately represented the grasping positions of the objects.
\end{abstract}

\section{INTRODUCTION}
Automatic object picking often requires a learning method that collects the training data online during the learning process. 
Such a method autonomously tries the predicted results, collects the data (trial samples), and evaluates
them. 
If the trial samples are successful, the method adds them to the dataset and returns to the training process using a deep neural network (DNN). 
The DNN automatically selects the feature vectors that accurately represent the desired output from a large training dataset \cite{r01}\cite{r02}. 
However, manually preparing the dataset is an expensive task. 
Online self-supervised learning reduces the manpower cost of preparing the training dataset for the DNN. 
\par
Unfortunately, the training data gathered by the autonomous learning method may degrade the training results. 
Learning methods such as reinforcement learning \cite{r2}\cite{r3} and sequential learning \cite{r4} collect the trial samples and provide teaching signals based on the success or failure of the trial. 
At this time, the collected data often lack a complete ground truth because the observable parts are limited. 
For example, the teaching signals exclude the information of the parts not sampled by an agent. 
In regular DNN training, the teaching signals are carefully evaluated by humans. 
Hence, a learning algorithm or evaluation method for the weakly labeled data acquired during the trial process is necessitated. 
Here, we address two problems in online self-supervised learning of object picking (Fig.1). 
\par

\setlength\textfloatsep{5pt}
\begin{figure}[t]
  \centering
  \includegraphics[width=7.6cm]{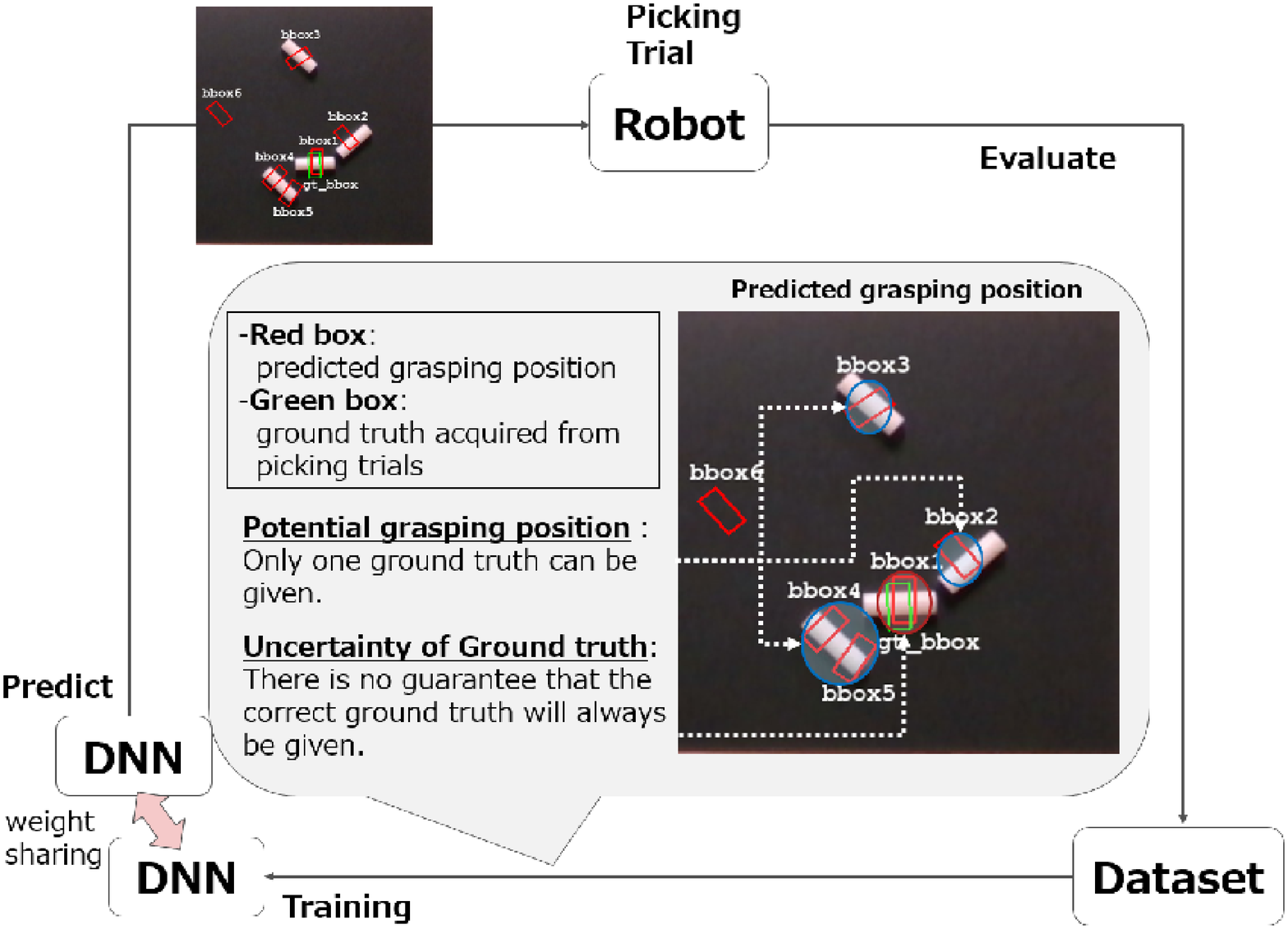}
  \caption{
  Online self-supervised learning for the object picking task. 
  The teaching signals in the training data are parial and uncertain. 
  }
\end{figure}

First, the uncertain ground truth of the trial samples may cause an undesirable training result. 
Recently, trial methods for gathering samples \cite{r5}\cite{r7} and the training algorithms of self-learning \cite{r8}\cite{r10} have been actively studied and improved. 
However, these studies do not optimize the individual goal of each trial sample; rather, they optimize the long-term rewards through reinforcement learning. 
The main objective of many robotic researches is to optimize the trajectory based on the success or failure of a task; an arbitrary target of the object operation is rarely given \cite{r2}. 
In most cases, the success condition of an object-picking task is given as a binary value indicating whether the robot grasped or missed the target object. 
The binary teaching signals quickly converge the learning process, but may include uncertainties that trap the training in a local solution. 
Consequently, the robot learns the grasp positions regardless of the state of the grasped object. 
This occurs because when all data are treated equally during the training process, the data acquired in the early training stages are preferentially learned over the later data. 
One solution is to weight each trial sample by the success degree of the sample, but the success estimation requires a detailed image processing of each object, which is extremely costly \cite{r11}. 
\par
Another problem arises when part of the ground truth is omitted from the trial sample. 
The missing data may decrease the prediction accuracy of the object picking. 
For instance, when the picking task involves multiple objects in the working area, the teaching signal is provided with only one trial result among the predicted grasping-position candidates; the teaching signals of the untried candidates are unknown. 
As the working environment may change after a trial, a retrial in the same situation is difficult. 
To compensate the missing information, researchers have proposed weakly-supervised learning \cite{r12}\cite{r13}. 
However, few methods have addressed the detection of multi-labeled grasping positions. 
\par
When the ground truth of the trial samples is uncertain, we evaluate the success degree of the trial samples as the teaching signal. 
The learning-based method is promising for evaluating trial samples without pre-design. 
DNN can vectorize the features of raw images. 
We compute the distance between the features represented in the DNN as the grasping score. 
As the distance is a continuous value, it provides more sufficient feedback than discrete values. 
In addition, our learning-based labeling method requires no detailed design in advance, so is easily applied to other objects. 
\par
Our method adjusts the feedback based on the grasping score and the potential grasping positions. 
Here the potential grasping positions define the correct position not attached to the ground truth in the trial sample. 
We also introduce a coefficient for controlling the loss for parameter optimization. 
This solution also removes the training instability caused by the order of data acquisition. 
The proposed online self-supervised learning method labels the training samples and applies appropriate feedback. 
These functionalities are realized by two DNNs: one that evaluates the trial samples with metric learning and provides teaching signals to the training dataset, and another that detects the optimum grasping position while adjusting the feedback amount. 
\par

\section{Related Works}
To ensure that the robot can perform tasks, the designed framework must evaluate sample trials and reproduce them during training. 
This is important because the trial samples acquired during self-supervised learning may contain unintended training data. 
\par
Deep reinforcement learning (DRL) is a precise learning method often applied in robotic picking tasks \cite{r14}. 
This method optimizes the trajectory policy based on pre-designed rewards. 
Robots governed by DRL perform tasks with sufficient accuracy, but the reward design is very sensitive. 
Even when the detailed rewards are designed in advance, the training may fall into a local solution unintended by the experimenter. 
For this reason, recent DRL methods are not targeted at cases requiring fine reward changes for each object. 
Designing the constraints on the optimum grasping positions of individual objects is a difficult task, and requires an excessively long training time \cite{r2}. 
In the method proposed in \cite{r3}, the relationship between the object grasping states and the image is learned by providing the image as the final goal. 
However, the design and embedding of the optimum grasping position, and the degree of the task success, are not considered in this model. 
\par
In simple tasks not requiring a long-term reward, stable training can be achieved by online self-supervised learning. 
In this circumstance, online self-supervised learning can be regarded as a simple training method requiring only one successful trial sample, as extreme training failures are rare. 
To detect the grasping position of an object, the authors of \cite{r99-3} and \cite{r16} applied a neural network with a convolutional layer. 
In \cite{r16}, the grasping probability of the surrounding pixels in the trial area is learned by treating the output of each pixel as a probability. 
Therefore, the model can adjust the feedback of the pixel regions not tried by the robot. 
However, these feedbacks are based on preliminarily designed distributions, and the feedback of a training dataset with an uncertain ground truth is not considered. 
\par
In other researches, the training uses the feature vectors obtained in the feature space of the DNN. 
Metric learning with a Siamese architecture \cite{r17} has been applied to similarity comparisons between two datasets \cite{r18}\cite{r19}. 
In a face-recognition study, the authors of \cite{r20} based the training error on the distance between feature vectors mapped to internal state of DNN, enabling face recognition with variations. 
As shown in these studies, metric learning is a promising approach for evaluating unknown data when the training data are sparse \cite{r21}\cite{r22}. 
Also, two DNNs have been combined in a behavior-learning framework using image feature vectors \cite{r24}\cite{r25}. 
However, neither of these methods evaluates the training data. 
Therefore, they are inapplicable to cases of sequentially collected trial samples. 
\par
The main contribution of our method is the evaluation of the success degree of a trial sample. 
The success is measured as the distance of the trial from the optimum grasping position in the feature space. 
Furthermore, a robot guided by the proposed method can precisely perform object-picking tasks when the grasping goal is designed from only a few images. 
\par
This study adopts the Siamese architecture \cite{r17} for online self-supervised learning. 
The model trains the similarity of the trial samples by metric learning. 
The teaching signal is based on the distance between two trial samples in the DNN feature space, which is defined in the Siamese architecture. 
As the teaching signal of the metric learning is the similarity measure between two input data of the DNN, the method is valid even when the ground-truth data are sparse, and is compatible with an autonomous learning method that always acquires new trial samples. 
\par
The proposed method also incorporates two types of feedback control; one that enlarges the feedback of trial samples with high similarity to the ground truth, and another that reduces the feedback of the grasping candidates estimated as potential grasping positions. 
These two kinds of feedbacks are designed to resolve the two problems described in Section. I. 
With these functionalities, only the good trial samples that guide the robot toward the optimum grasping position are preserved for training. 
\par

\section{METHOD}
We propose an online self-supervised learning method that detects a robot’s optimum grasping position. 
The proposed method (Fig.2) uses two types of DNNs: (a) single shot multibox detector (SSD) that detects multiple grasping positions of the object, and (b) Siamese networks (SNs) that evaluate the success degree of the trial samples. 
Our main ideas are summarized below: 
\begin{itemize}
  \item Adjust the amounts of both kinds of feedback for the detected grasping-position candidates or the background.
  \item Evaluate the trial samples by their similarities to other trial samples in the feature space.
\end{itemize}
\par
To realize these ideas, we estimate the distances between pairs of trial samples in the feature space using metric learning. 
This computation is performed by the SNs. 
After training the similarity of two trial samples, the distance represents the deviation of the trial sample from the optimum grasping position. 
The SNs embed the semantically similar examples close together using a distance-based loss function. 
In addition, when training the SSD to detect the grasping positions, the model adjusts the training error of each prediction result with the grasping score and infers the number of potential grasping positions. 
The SSD preferentially trains the trial samples near the optimum grasping position. 
\par
The trial samples with their grasping scores evaluated by the SNs are added to the training dataset. 
The SNs and SSD are trained together. 
The details of the two DNNs are described in subsections A and B, and the training procedures of the SNs in online self-supervised learning are described in subsection C. 
\par

\setlength\textfloatsep{5pt}
\begin{figure*}[t]
  \centering
  \includegraphics[width=16.5cm]{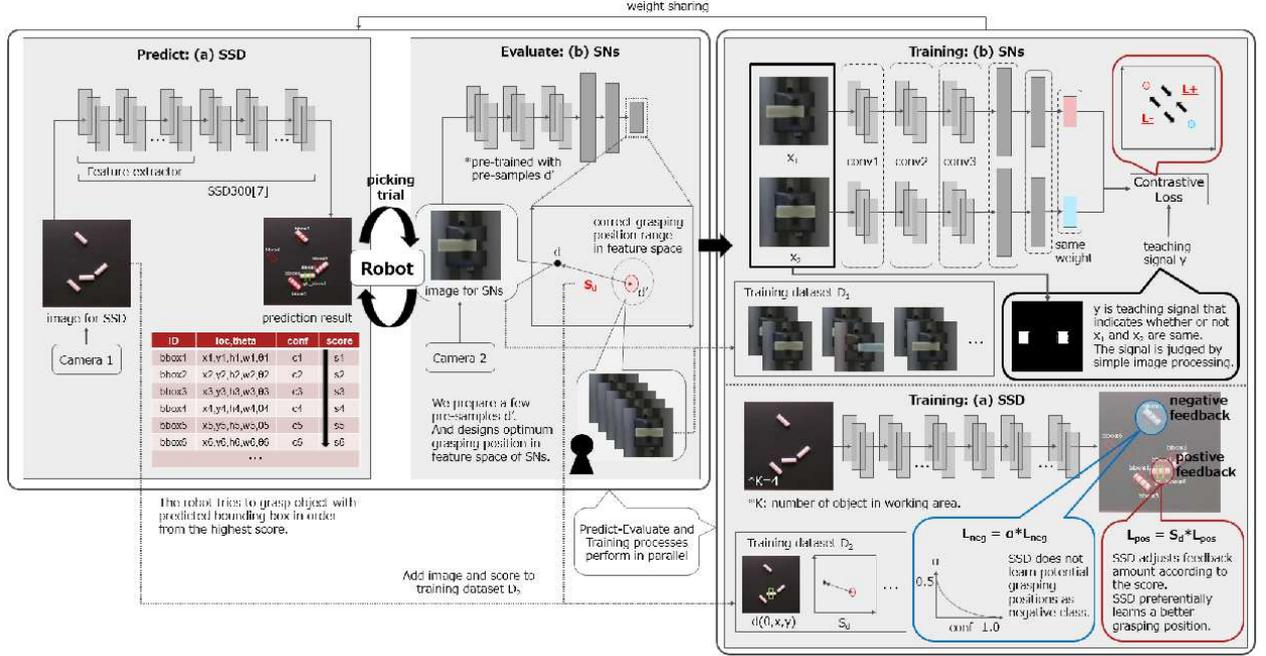}
  \caption{
  Overview of our method with two-DNN method. 
  The prediction–evaluation and training processes are performed in parallel. 
  The SNs embed the similar trial samples nearby the optimum grasping position, and the dissimilar trial samples further away in the feature space. 
  The SSD trains the grasping positions aided by two kind of adjustable feedbacks: the positive feedback based on the grasping score evaluated by the distance in the
feature space of the SNs, and the negative feedback  determined by the potential grasping position. 
  }
\end{figure*}

\subsection{Single Shot Multibox Detector}
To train the grasping position of the object in the proposed method, we implemented and extended the SSD \cite{r99-1}. 
The SSD detects multiple bounding boxes using variously sized convolution layers. 
By directly predicting the bounding box of the grasping position in the raw image, the method enables speedy detection. 
The bounding boxes are classified into the background (negative) class or the grasping-position (positive) class. 
The SSD is optimized to minimize the training errors from positive and negative feedback. 
In this study, the model is extended to object picking tasks as follows. 
\par
{\it 1) Detection of the grasping position:}
The SSD in our method trains and predicts the rotation angle $\theta$ and the grasping score $s$ in each bounding box. 
Each bounding box has two object class, "object" or "background". 
By applying the rotation angle to the bounding box, we can predict the grasping area on the two-dimensional image. 
The grasping score, which predicts the optimality of a grasping-position candidate, is the criterion for choosing the grasping position in the trial process and the test phase. 
The teaching signal of the grasping score is evaluated by the SNs. 
\par
{\it 2) Adjustment of feedback amount:}
The second SSD extension enables feedback adjustment the grasping-position candidate during training:
\begin{itemize}
  \item Adjust the negative feedback with the potential grasping positions.
  \item Adjust the positive feedback with the grasping score. 
\end{itemize}
As described in Section I, the training data are sequentially inserted into the online self-supervised learning. 
Therefore, the teaching signals are often insufficient for training the optimum grasping position. 
When training the background class, only one ground truth is provided for each image of the trial result, so some potential grasping positions may be incorrectly trained as the background class. 
The given ground truth may also contain subtle errors. 
\par
To handle the above problems, we incorporate a mechanism that controls the feedback of each bounding box. 
By adjusting the negative and positive feedback respectively, the model promotes the training of good grasping positions and disfavors training the bad ones. 
The loss function of SSD is calculated as follows:
\begin{eqnarray}
L = \sum \left( SL_{pos} + \sum_{k} \alpha_{k} L_{neg, k} \right)
\end{eqnarray}
\begin{align}
\alpha_k &= 
\left \{
\begin{array}{lll}
\cfrac{1}{2} \,\, (1.0 - conf_k)^2, \quad if \,\, k\,\leq\,K,\\
\\[-4.0mm]
1.0, \quad\,\,\,\,\,\,\,\,\,\,\,\,\,\,\,\,\,\,\,\,\,\,\,\,\,\,\,\,\,\,\,\,\,\,\, otherwise
\end{array}
\right.
\end{align}
where $L_{pos}$ is the positive feedback for the grasping position candidates, and $L_{neg}$ is the negative feedback for background candidates. 
$S$ represents the grasping score given by the SNs. 
$conf$ is the classification probability of the bounding box belonging to the background class. 
The coefficient $\alpha_{k}$ adjusts the negative feedback of the rectangle with the $k$-th highest $conf$ value.  
\par
Note that when adjusted by $\alpha_{k}$, the negative feedback reduces as $conf_{k}$ increases over the range of $K$, where $K$ is the estimated number of potential grasping positions. 
$K$ actually defines the number of objects in the acquired image during the picking task and is assumed to be obtainable by outside processing. 
\par
Meanwhile, the positive feedback is adjusted by multiplying the feedback term by the grasping score $S$. 
The grasping score increases when the bounding box encloses an appropriate grasping position. 
Under this feedback control, the optimum grasping positions are preferentially trained. 
\par

\subsection{Siamese Networks}
To adjust the SSD feedback, we require criteria for evaluating the grasping score of the acquired trial samples. 
As described in Section I, a learning-based approach can assign different evaluation indicators to various objects. 
In this study, we apply metric learning and evaluate the grasping score using the distance in the feature space of the DNN. 
In metric learning, the similarity between pairs of training data is embedded in the feature space. 
Hence, the differences between the trial samples can be represented by continuous values. 
Also, as the difference between two trial samples is a teaching signal in the proposed method, a detailed labeling of all trial samples is unnecessary, and the optimum grasping position of an individual object is easily designed. 
\par
The SNs in our method evaluate the success degree of each acquired trial sample. 
Owing to their Siamese architecture \cite{r17}, these DNNs train well even when the training datasets per class are extremely small \cite{r21}\cite{r22}. 
The two SNs in our network design share the same weights. 
Each SN receives a different data input, and the output vectors of the two SNs are compared. 
Because the SNs process a combination of two input data, the training dataset becomes larger than in supervised training. 
This approach is suitable for online self-supervised learning when the volume of acquired data during the training process is limited. 
\par
In this study, the SNs convert the two inputs into feature vectors with entries ranging from $-1$ to $1$. 
The model then trains the distance between the two vectors based on the teaching signal $y$, which indicates whether the inputs are the same or different. 
The contrastive loss function of the SNs for optimizing the distance between the two input data $x_{1}$ and $x_{2}$ is calculated as follows: 
\begin{eqnarray}
L = \sum \cfrac{1}{2} ((1 - y) * L_{+} +  y * L_{-} )
\end{eqnarray}
\begin{eqnarray}
L_{+} = D(x_{1}, x_{2})^2
\end{eqnarray}
\begin{align}
L_{-} &= 
\left \{
\begin{array}{lll}
(margin - D(x_{1}, x_{2})^2), \quad if \,\, L_{-}\,\geq\,0,\\
\\[-4.0mm]
0, \quad\,\,\,\,\,\,\,\,\,\,\,\,\,\,\,\,\,\,\,\,\,\,\,\,\,\,\,\,\,\,\,\,\,\,\,\,\,\,\,\,\,\,\,\,\,\,\,\,\,\,\,\,\,\,\,\,\, otherwise
\end{array}
\right.
\end{align}
where, $L_{+}$ and $L_{-}$ indicates similar loss and dissimilar loss, respectively, $D$ is the Euclidean distance between the input data, and $margin$ is a coefficient for adjusting the contrastive loss. 
\par
During training, a pair of output vectors either converges (if similar) or diverges (if dissimilar). 
Therefore, the feature space of the trained SNs represents the differences between pairs of trial samples. 
The feature vectors nearby the optimum grasping position represent the useful trial samples for training. 
The proposed method assigns the grasping score (distance between the trial sample and the optimum grasping position) to the training dataset of the SSD. 
The following subsection describes the incorporation of SNs into the online self-supervised learning and estimates the grasping score $S$. 
\par

\subsection{Self-Supervised Learning with Metric Learning} 
Before the SNs can assign a grasping score to the trial samples, they must calculate the distance between the sample and the feature vector of the optimum grasping position. 
For this purpose, we prepare some pre-samples and train the SNs in advance. 
Pre-samples are designed with human hands to fit specified grasping-position. 
The pre-samples are images of the optimum grasping position and a slightly sub-optimal grasping position. 
During the pre-training, the optimum and sub-optimum grasping positions are labeled, and a threshold value for the teaching signals is also decided. 
At this time, a threshold value for the teaching signals is also decided. 
As only the pre-samples are required for designing the optimum grasping position, our method is flexibly applicable to various objects at low introduction cost. 
\par
During online self-supervised learning, the SNs compare the trial sample $d_{t_1}$ acquired in the $t_{1}$th trial with the trial sample $d_{t_2}$ acquired in the $t_{2}$th trial. 
Trial samples that completely failed the grasping operation are excluded from this step. 
As the SNs optimize the similarity between two input data, individual labels for the trial samples are not needed. 
In this study, the teaching signal $y$ is determined by calculating the difference between each pixel of the two images. 
If the number of different pixels exceeds the threshold value, the two input images are deemed to be different. 
\par
The grasping score $S_{d}$ of trial sample $d$ defines the distance between $d$ and the optimum grasping-position group. 
The grasping score $S_{d}$ d is defined as follows: 
\begin{eqnarray}
S_{d} = norm\left(
\sqrt{\sum_{i} (d_{i}-\overline{d^{\prime}_{i}})^2}
\right)
\end{eqnarray}
Here, $i$ represents an individual element of $d$, and $\overline{d^{\prime}}$ represents the average distance over the pre-samples $d^{\prime}$ . 
The $norm$ function normalizes the processing so that $S_{d}$ is large when the trial sample approaches the optimum grasping position. 
Because the SNs train the trials in parallel, the latter portion of the training is labeled more accurately than the former portion. 
To correct this performance drift, we reset the grasping score of the SSD dataset at regular intervals. 
The parameters of the network are not initialized but the stochastic gradients by the optimizer are computed from the updated data-set after the interval. 
This resetting stabilizes the training of the whole method. 
\par

\section{Experiment}
To confirm whether training by the proposed method captures the success degree of the trial without designing rigorous evaluation criteria beforehand, we executed cylinder pickup by a robot arm \cite{r26} (Fig.3). 
The cylindrical object is suitable for evaluating the basic effectiveness of the proposed method. 
Especially, the deviation of the grasping position of a simple shape such as a cylinder is easily determined by image processing. 
Hence, the experiment is easily carried out under several success conditions. 
Five cylinders were set at random positions on the working area in front of the robot. 
The robot picked up the cylinders in sequence, and the proposed method trained the pickups in parallel. 
After the cylinder-picking task, we applied our method to some elongated objects (pen, scissors, ladle, and rice paddle). 
\par

\subsection{Design of the Picking Task}
Algorithm 1 implements the online self-supervised learning and specifies its parameters. 
First, we collect pre-samples $d^{\prime}$ to design the optimum grasping position and pre-train the SNs with them. 
The picking experiment is then started. 
During the experiment, the SSD predicts the grasping-position candidates of the object in the image captured by Camera 1. 
The robot tries the grasping positions in order sequence, starting from $d_{pred,s_{j}}$ with the highest predicted grasping score $s$. 
At this time, if the trials have been performed $N_{trial}$ or more times, the robot attempts random positions $d_{rand}$ within the working area. 
The movement extent in the z direction and the opening width of the gripper are fixed. 
\par
The acquired trial sample $d$ was judged as successful or unsuccessful by the gripper opening and closing width. 
When the trial was successful, an image was acquired from Camera 2 affixed beside the robot. 
The acquired image was input to the SNs and mapped onto the feature space. 
The grasping score $S_{d}$ was calculated as the distance of the trial sample from the optimum grasping-position group in the feature space. 
The trial sample with its grasping score is was added to the datasets $D_{1}$ and $D_{2}$. 
\par
After the trial, if no object appeared in the working area, the objects were reset in the working space, and the SNs recalculated the grasping score of the training dataset. 
These processes were repeated a predetermined number of times $N_ {env}$. 
\par

\begin{figure}[t]
  \centering
  \includegraphics[width=7.5cm]{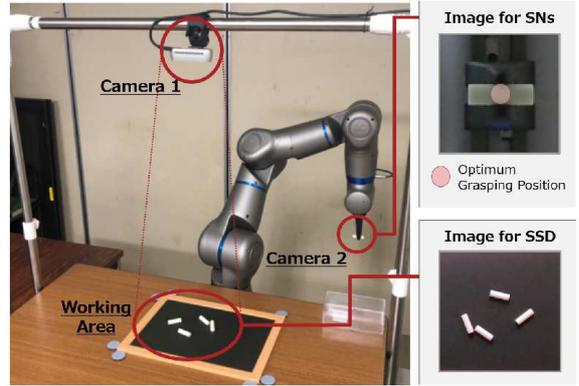}
  \caption{
  Cylinder-picking task by the robot arm. 
  Two cameras capture the images input to the DNNs. 
  The red circle in the SN image encloses the optimum grasping position. 
  }
\end{figure}

\begin{algorithm}[t]
\caption{Online Self-Supervised Learning Process}
\begin{algorithmic}[1]
\STATE add $d^{\prime}$ to $D_{1}$ and pre-training SNs
\STATE start training SSD and SNs
\FOR{$(n\leq\,N_{env})$}
    \STATE $i, j \leftarrow N_{obj}, 0$
    \WHILE{($i > 0$)}
        \STATE capture image1 and predict $d_{pred}$ by SSD
        \IF{$j < N_{trial}$}
            \STATE $d(x,y,\theta) \leftarrow d_{pred,s_{j}}(x,y,\theta)$
            \STATE $j \leftarrow j + 1$
        \ELSE
            \STATE $d(x,y,\theta) \leftarrow d_{rand}(x,y,\theta)$
        \ENDIF
    \IF{success to pick up}
        \STATE capture image2 and evaluate $S$ by SNs
        \STATE add image1 and $d(x,y,\theta), S$ to $D_{1}$
        \STATE add image2 and $y$ to $D_{2}$
        \STATE $i, j \leftarrow i- 1, 0$
    \ENDIF
    \ENDWHILE
    \STATE set $S$ of trial samples in $D_{1}$ again
\ENDFOR
\end{algorithmic}
\textbf{Parameters:} $d^{\prime}$=30, $N_{env}$=50, $N_{trial}$=5, $N_{obj}$=5
\end{algorithm}

\subsection{Training and Model Setup}
In the experiment, two types of images were captured as input. 
The images input to the SSD were acquired from Camera 1 set at the top of the working area, and inputs to the SNs were acquired from Camera 2 set beside the robot. 
The images from both cameras were sized $300\times 300 \times3$ pxls, and were augmented to increase the robustness of DNN. 
\par
To learn the object picking task, we must set the parameters of the proposed method. 
The SNs are composed of two convolutional layers and three fully-connected layers. 
The $margin$ of the contrastive loss was 1.0. 
The grasping position was detected by the SSD300, which predicted the $loc, conf, \theta$ and $s$ in each rectangular area. 
Both DNNs were  optimized by Adam. 
The parameters of the SNs and SSD are listed in Table I. 
The chosen parameters were those yielding the best results in a trial-and-error process. 
In addition, the training was performed on random seeds with five values of initial weighting schemes. 
\par

\begin{table}[thbp]
  \centering
  \begin{tabular}{c|c}
    \multicolumn{2}{c}{TABLE I: Structures of the networks} \\
    \hline
    Network         & Dims \\
    \hline \hline
    SNs*    & input@3chs - conv@20chs - conv@50chs - \\ 
            & full@500 - full@10 - full@2    \\ 
    \hline
    SSD     & input@3chs - SSD300 \cite{r99-1} -  \\ 
            & output($loc,conf,\theta,s$) \\ 
    \hline 
  \end{tabular}
  \begin{center}
  * all conv filters are kernel size 5, stride 1, padding 1 \\
  \end{center}
\end{table}

\vspace{-2.0mm}
\subsection{Evaluation Setup}
If the robot can grasp the object at an appropriate position, the trained model is considered to have good detecting ability. 
In this experiment, the grasping performance was evaluated after each iteration. 
The model predicted the grasping positions of one to five objects set at random positions on the working area. 
The robot executed 20 trials of each setup (a total of 100 trials). 
The success rate of our method was compared with that of a baseline method under several grasping conditions. 
\par
Moreover, to confirm whether our method labels the trial samples using appropriate evaluation criteria, we verified the feature space of the SNs. 
If the distance between each trial sample and the optimum grasping position corresponds to the real grasping position, then the feature space of the SNs is appropriate for evaluating the success degree of the trial. 
For this purpose, we visualized the feature space of the SNs while the robot grasped the cylinder in various positions. 
\par

\section{Results and Discussion}

\begin{figure}[t]
  \centering
  \includegraphics[width=7.5cm]{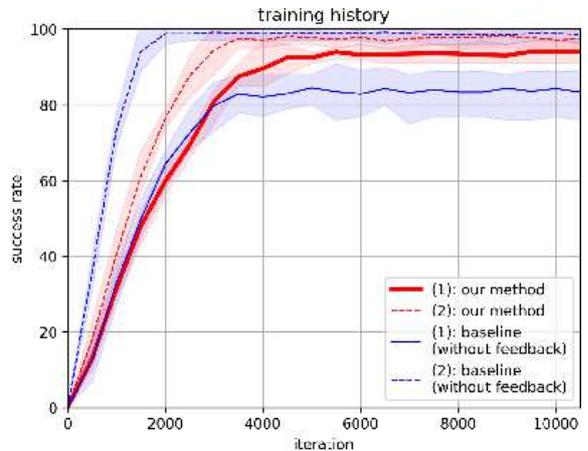}
  \caption{
  Success rate of the cylinder-picking task. 
  The blue and red curves are the results of the baseline and proposed methods, respectively. 
  The solid and broken lines indicate conditions (1) (grasping the cylinder within $5^\circ$ angular error and 1.0 cm positional error) and (2) (grasping the cylinder from any position), respectively.
  }
\end{figure}

\subsection{Success Rate of the Picking Task}
We first verified the performance of the our method in a picking test. 
We compared the baseline and proposed method. 
In the baseline method, the teaching signals of the trial samples are given as binary signals (success or failure for grasping an object). 
Therefore, the $\alpha_{k}$ and grasping score $S$ always equal 1.0. 
Fig.4 plots the success rates of the picking task during training. 
The robot was required to grasp the cylinder under two conditions. 
\begin{description}
  \item[(1)] Grasping within an angular error of $5^\circ$ and a positional error of 1.0 cm.
  \item[(2)] Successful grasping at any grasping position.
\end{description}
\par
After 10000 iterations, the success rates of the baseline and proposed methods were 84.5\% and 94.5\% under condition (1), respectively. 
Hence, our method improved the accuracy of training. 
In addition, the accuracies of the baseline method and our proposed method were almost equal under the easy condition (2), confirming that our method can successfully train when the teaching signals are insufficient. 
\par
To confirm the possibility of training for any optimum grasping position designed by the experimenter, we trained the grasping position of the left or right side of the cylinder. 
The success rates of grasping from the left and right under condition (1) were 94.2\% and 95.4\%, respectively. 
The accuracies of both were almost the same as the result in Fig.4, confirming that our method can learn the optimum grasping position from only a small number of pre-samples given beforehand.
\par
We also applied our method to some elongated objects on the desk (Table II). 
In all cases, the method was able to detect was better grasping-positions than baseline. 
The success rate of ladle and rice paddle ware low because the posture of the objects are not stable. 
Since the model trained from the raw image with data augmentation, it has the robustness and can predict complicated shapes. 
\par

\begin{table}[thbp]
  \centering
  \begin{tabular}{c|cccc}
    \multicolumn{5}{c}{TABLE II: Success rate under condition (1)} \\
    \hline
               & cylinder & pen & scissors & ladle and \\
               &          &     &          & rice paddle \\
    \hline     \hline
    baseline   & 84.5\% & 84.2\% & 75.7\% & 62.5\% \\ 
    \hline
    our method & 94.5\% & 93.8\% & 87.4\% & 71.2\%  \\ 
    \hline 
  \end{tabular}
\end{table}

\begin{figure}[htpb]
  \centering
    \begin{tabular}{c}
      \begin{minipage}{0.40\hsize}
        \centering
          \includegraphics[width=3.5cm]{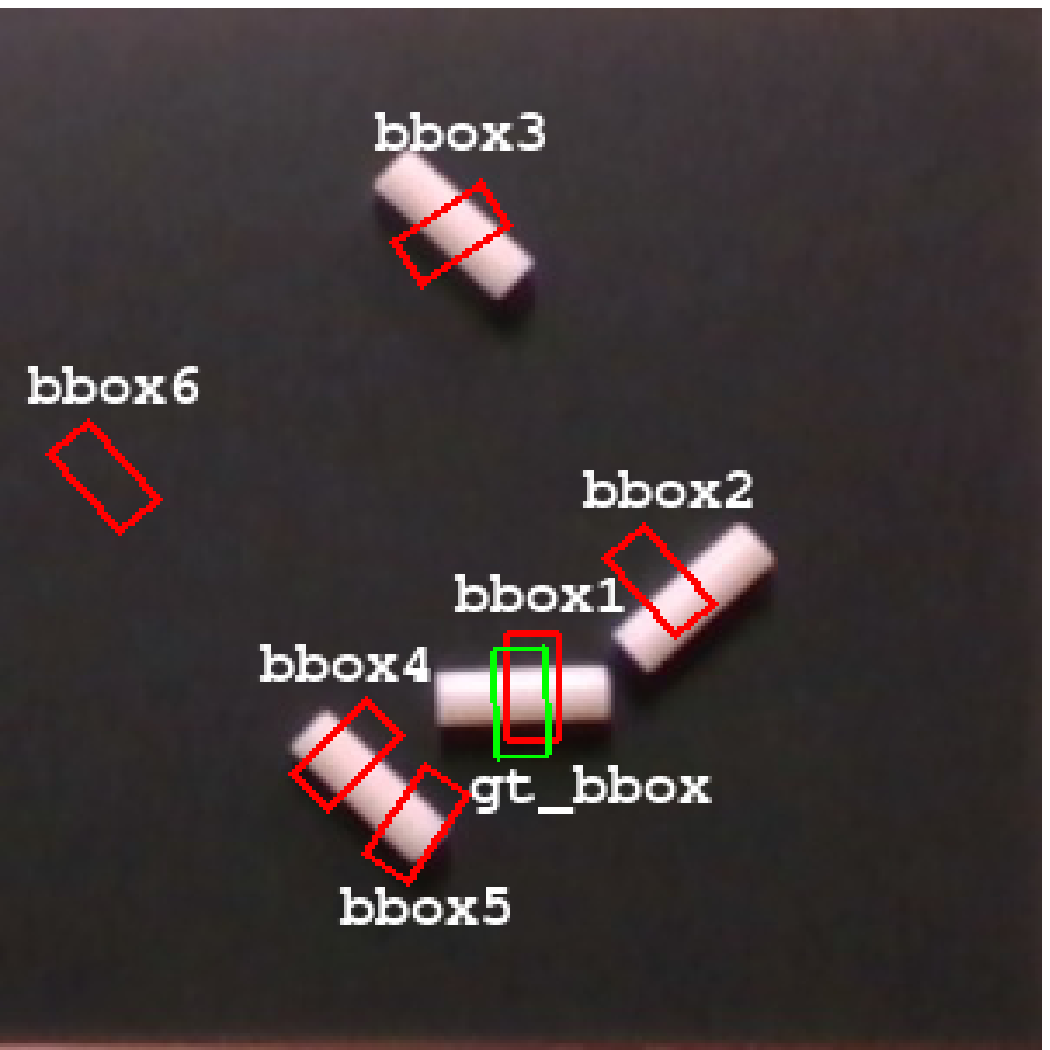}
      \end{minipage}
      \begin{minipage}{0.50\hsize}
        \centering
          \includegraphics[width=3.5cm]{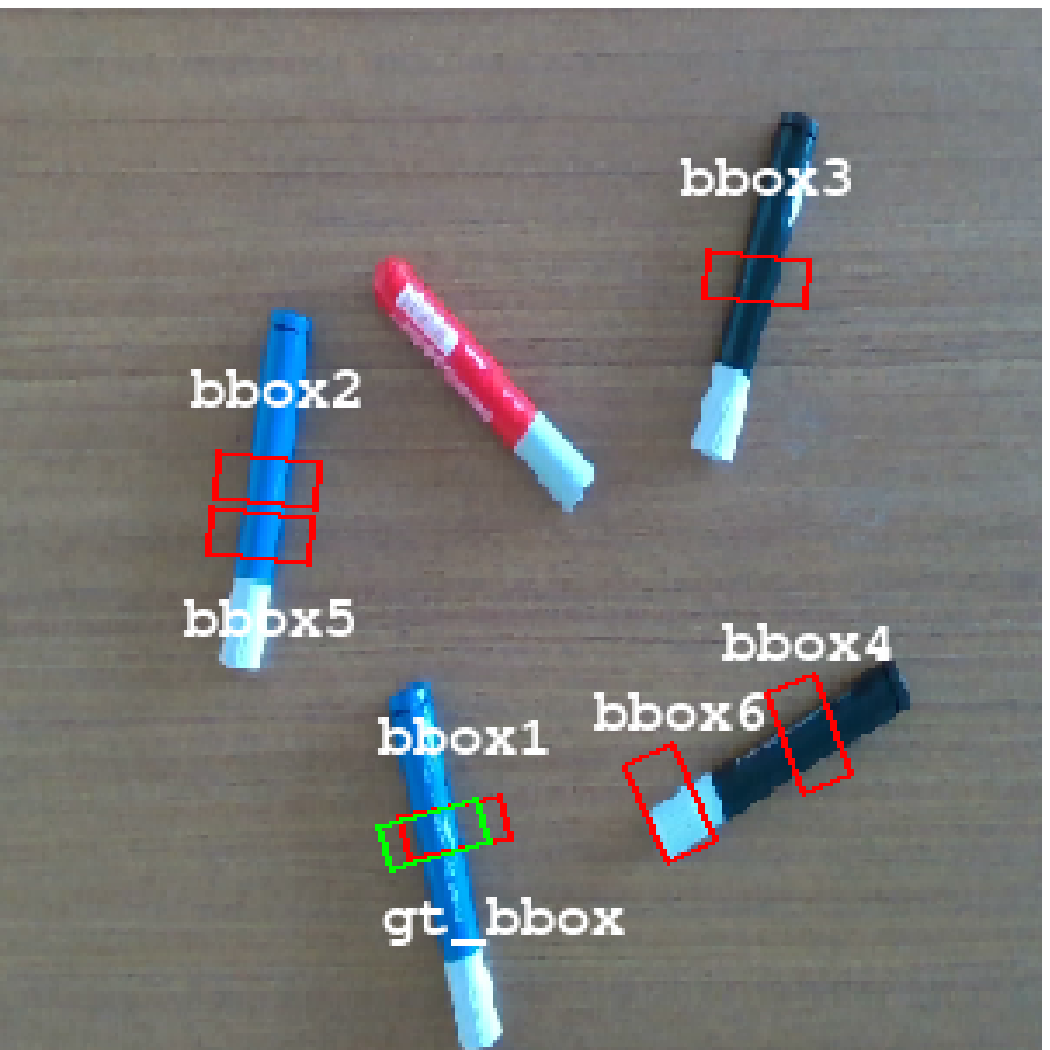}
      \end{minipage} \\

      \begin{minipage}{0.01\hsize}
        \vspace{-8.0mm}
      \end{minipage} \\

      \begin{minipage}{0.40\hsize}
        \centering
          \includegraphics[width=3.5cm]{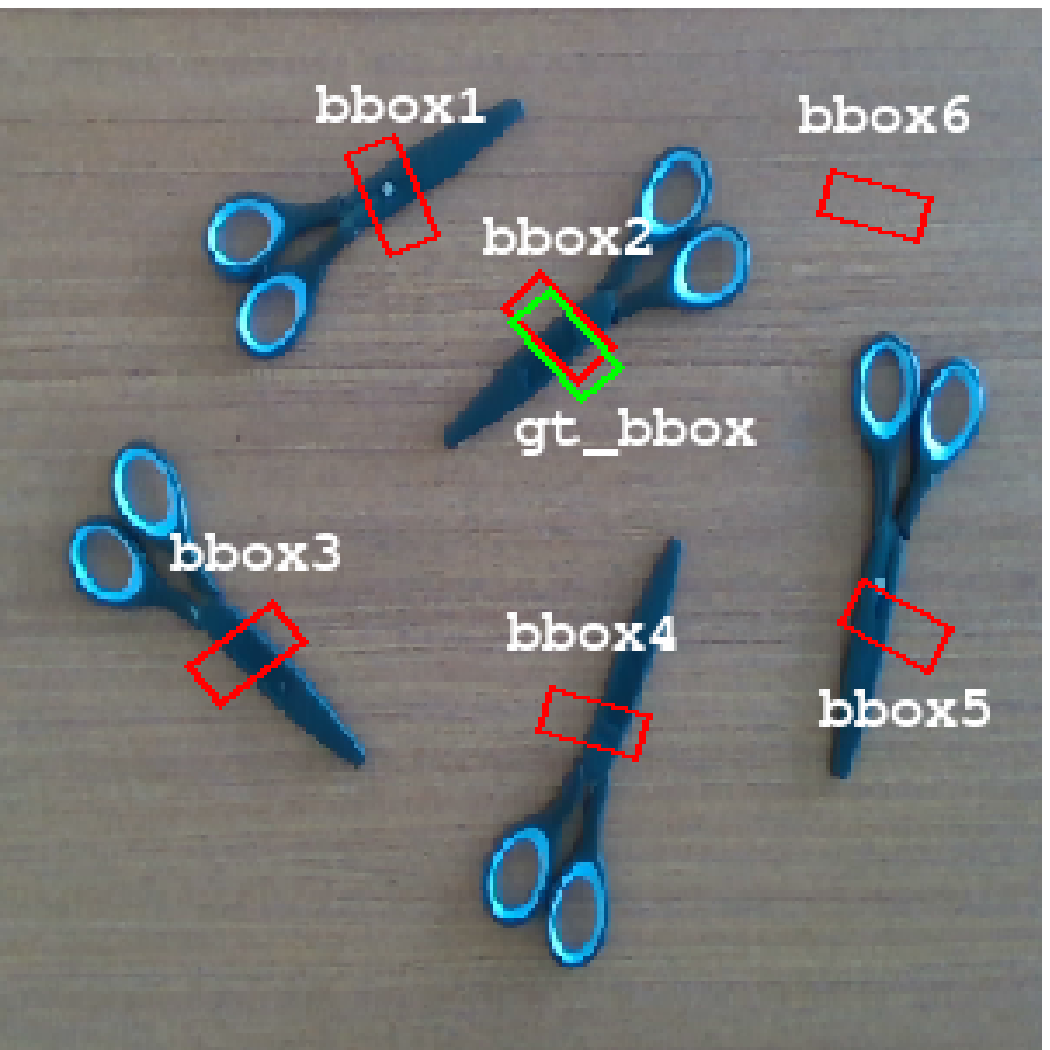}
      \end{minipage}
      \begin{minipage}{0.50\hsize}
        \centering
          \includegraphics[width=3.5cm]{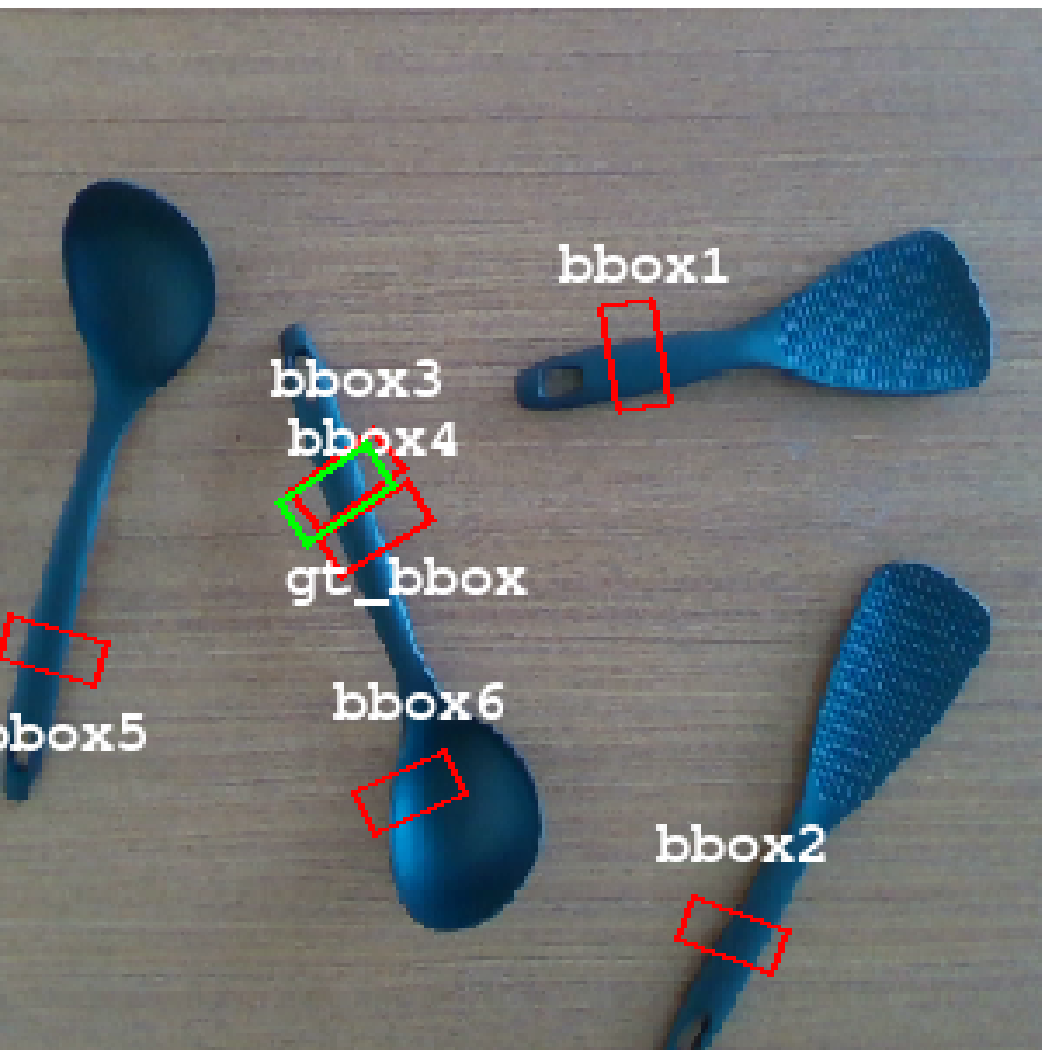}
      \end{minipage}
    \end{tabular}
  \caption{
    Example of detecting the grasping position while training the SSD. 
    The red and green boxes indicate the SSD prediction results and the ground truth given in the trial result, respectively. 
    }
\end{figure}

\subsection{Feedback during Training SSD}
To confirm the feedback function of our method, we visualized the prediction results of the SSD during the training process. 
In the Fig.5, bounding boxes $bbox$ are numbered in descending order of their $conf$ value. 
Boxes $bbox2$ to $bbox5$ are considered as the potential grasping positions in Eq.(2), so their errors are small. 
This result indicates bad trial samples ware ignored by individually adjusting the feedback of each training dataset. 
\par
However, the feedback range $K$ depends on the shape of the grasped object. 
When the object can be grasped from multiple positions, the number of potential grasping positions becomes more important. 
In such cases, the number of objects in the working area can be inferring by inferred by a DNN or image processing.
\par

\subsection{Feature Space in Siamese Networks}
To verify whether the SNs assign an appropriate grasping score to the trial samples, we visualized the feature space of the SNs after inputting some sampled data from the test dataset (Fig.6). 
The robot was required to grasp the center, left, or right position of the cylinder. 
In the center grasping case (Group 1), the points were collected around the optimum grasping position group. 
Moreover, the distance in the feature space corresponded to the actual grasping position of the cylinder. 
As examples, Fig.6 displays the images of several holding positions input to the SNs. 
When the robot grasped the cylinder slightly to the left of its optimum grasping position (Group 2), the points plotted close to those of Group 1. 
On the other hand, when the robot grasped the cylinder at the distant right of its optimal position, the points plotted far from those of Group 1. 
This result indicates that the SNs appropriately evaluated the success degree of the trial samples. 
\par
During the test phase, the robot never grasped the cylinder at a position far from its optimum grasping position, because in such a circumstance, the object state on the input image greatly deviated from the successful object state. 
When the output vector greatly departed from the feature group of the optimum grasping position, the grasping score assigned to the trial sample was considerably reduced. 
\par
Although there are some works of self-supervised learning with estimating grasping rectangles for various objects\cite{r4}, they are not focused on grasping position that fit for object or task. 
Our approach can predict optimum grasping position to fit any robot picking task. 
\par

\begin{figure}[t]
  \centering
  \includegraphics[width=7.5cm]{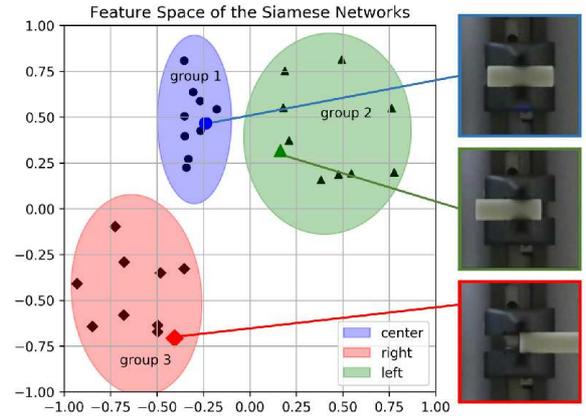}
  \caption{
  Some trial samples in the feature space of SNs. 
  Within the blue space, the robot grasped the center position of the cylinder. 
  In the red and green spaces, it grasped the right and left sides of the cylinder, respectively. 
  }
\end{figure}

\subsection{Training Stability and Future Work}
In online self-supervised learning, the prediction and feedback of the grasping score effectively improve the stability of training. 
Generally, when the configuration of the dataset changes online, the training result depends on the order of the acquired samples. 
For example, if many bad samples are accumulated in the dataset during the early training stage, the model often predicts a bad grip position. 
This is because each trial sample is trained equally and assigned a uniform grasping score. 
\par
In our method, the feedback values of the trial sample with a high grasp score increase regardless of the acquisition order. 
The coefficient $S$ in Eq. (1) stabilizes the training process. 
The training converged in all training experiments performed in this study. 
In addition, the final success rate was less variable in our method than in the baseline method. 
However, the stability of our method largely depends on the pre-training results of the SNs. 
To resolve this problem, we must consider the number and quality of the pre-samples. 
\par
When handling multiple kinds of objects or multiple potential ground truths, our method have to extend model architecture. 
For example, the SNs evaluates each object at the output convolutional layer that replaced from the fully-connected layer. 
In this way, our method has the potential for expansion to the various task.

\section{Conclusion}
We proposed the method that detects the optimum grasping position of objects by online self-supervised learning with two types of DNNs. 
Our method embeds the grasping position of the object in the feature space of the SNs, enabling fluent feedback without a prior detail design. 
The SSD was trained by computing the distance of the sample from the optimum grasping position in the feature space. 
Its feedback was adjusted for the closeness between the potential and ground-truth grasping positions. 
The proposed online self-supervised learning was experimentally validated in a cylinder-picking task by a robotic arm. 
The detection and evaluation of the optimum grasping position was confirmed by visualizing the feature space of the SNs. 
As demonstrated in our results, the proposed method mapped the success degree of the trial samples on the feature space. 
Moreover, the robot performed the task more accurately with our method than the baseline method. 
\par
In future work, we hope to extend our method to more complex objects and different tasks.


\end{document}